\documentclass[runningheads]{llncs}
\usepackage[T1]{fontenc}
\usepackage[dvipdfmx]{graphicx}

\usepackage{subfigure}
\usepackage{multirow}
\usepackage[dvipdfmx]{color}

\begin{document}
%
\title{Learning from AI: An Interactive Learning Method Using a DNN Model Incorporating Expert Knowledge as a Teacher}
%
%

\author{Kohei Hattori\inst{1,2} \and
Tsubasa Hirakawa\inst{1,3} \and
Takayoshi Yamashita\inst{1,4} \and
Hironobu Fujiyoshi\inst{1,5} }

\authorrunning{H. Kohei et al.}
%
\institute{Chubu University, Kasugai Aichi, Japan \and 
\email{tr20011-6745@sti.chubu.ac.jp} \and
\email{hirakawa@mprg.cs.chubu.ac.jp} \and
\email{takayoshi@isc.chubu.ac.jp} \and
\email{fujiyoshi@isc.chubu.ac.jp}}
\maketitle              
\begin{abstract}
Visual explanation is an approach for visualizing the grounds of judgment by deep learning, and it is possible to visually interpret the grounds of a judgment for a certain input by visualizing an attention map. 
As for deep-learning models that output erroneous decision-making grounds, a method that incorporates expert human knowledge in the model via an attention map in a manner that improves explanatory power and recognition accuracy is proposed. 
In this study, based on a deep-learning model that incorporates the knowledge of experts, a method by which a learner “learns from AI” the grounds for its decisions is proposed. 
An “attention branch network” (ABN), which has been fine-tuned with attention maps modified by experts, is prepared as a teacher. 
By using an interactive editing tool for the fine-tuned ABN and attention maps, the learner learns by editing the attention maps and changing the inference results. 
By repeatedly editing the attention maps and making inferences so that the correct recognition results are output, the learner can acquire the grounds for the expert's judgments embedded in the ABN. 
The results of an evaluation experiment with subjects show that learning using the proposed method is more efficient than the conventional method. 

\end{abstract}
\section{Introduction}
Visual explanation is an approach that visually interprets the grounds of a judgment by visualizing the region of attention \cite{CAM,Grad-CAM,ABN} during inference by a convolutional neural network (CNN) \cite{AlexNet,VGG,ResNet,GoogLeNet,SENet}. 
As one visual-explanation method, an attention branch network (ABN) \cite{ABN} improves classification accuracy by introducing an attention mechanism that visualizes an attention map as a grounds for making decisions about classification results in an image-classification task and weights feature maps in regard to gaze regions. 

Although an ABN makes it possible to visually understand the grounds for decisions, if the number of training data for the target task is insufficient or if label noise is caused by annotation errors, two problems arise: (i) classification accuracy is reduced and (ii) and obtaining an attention map that provides the correct grounds for decisions becomes difficult. 
In such cases, the reliability of the attention map decreases, and it becomes difficult to understand the correct grounds for decisions. 
In response to these problems, a method for introducing expert knowledge into the network was proposed by Mitsuhara et al \cite{re_teach}. 
As for this method, the ABN is fine-tuned by using an attention map modified by an expert in image-classification tasks, and on the grounds of the modified attention map and attention mechanism of the ABN, visual explanation and recognition performance are improved.

Various studied have attempted to help people understand AI behavior and intervene in AI learning by introducing knowledge through such visual explanations. 
In contrast, this study aims to create a “human-in-the-loop” mechanism, by which people learn from AI by not only humans intervening in AI learning but also AI intervening in human learning. To achieve that aim, in this paper, we propose a method by which humans learn the grounds for their decisions from AI (i.e., they “learn from AI”) by using a deep-learning model (incorporating the knowledge of experts) as a teacher. 
As for the proposed method, the learner manually edits an attention map and uses it to perform inference of an ABN (incorporating an expert's knowledge) and check the results. 
The attention map is repeatedly and interactively edited so that the inference result of the ABN is correct and attains a high score. Humans can learn the grounds for the judgments of the expert embedded in the ABN through attention map. 
To implement this learning method an educational application was created. 
In an experiment, we used an educational application that we created to evaluate the proposed method. In particular, we evaluated subjects tasked with judging disease in fundus images. 
As a result of the experiment, we evaluated the learning effect of the proposed method and the educational effect of the created application.

The contributions of this study is summarized as follows:
\begin{itemize}
\item A method by which AI intervenes in human learning is proposed. In detail, the learner edits an attention map and checks changes in the inference results given by an ABN incorporating an expert's knowledge; as a result, the learner learns the grounds for the expert's judgment.
\item To enable effective learning from AI, an educational application by which a learner learns while interacting with AI through editing an attention map was created, and the effectiveness of using the app was verified.
\end{itemize}

\section{Related studies}
\begin{figure}[tb]
\centering
\includegraphics[width=12.2cm]{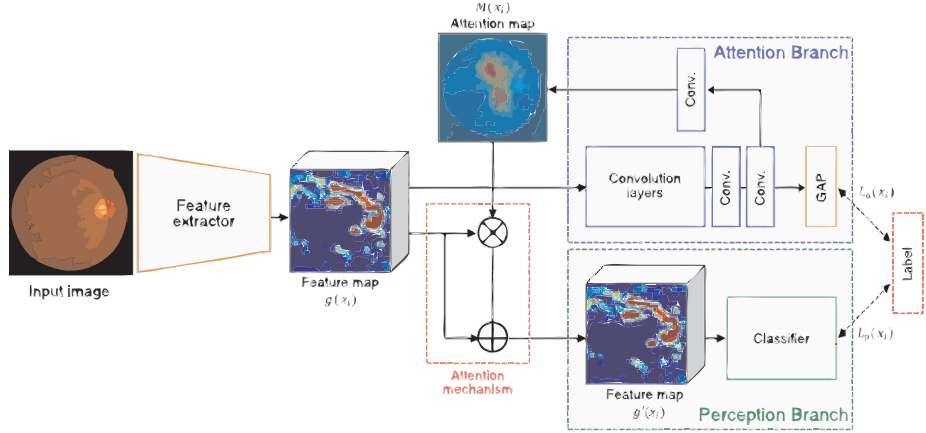}
\caption{Configuration of an attention branch network}
\label{fig:abn}
\end{figure}
Visual explanation is a method for visualizing the grounds for decisions of deep-learning models, including CNNs, by using an attention map. 
Typical visual-explanation methods include class activation mapping (CAM) \cite{CAM}, gradient-weighted CAM \cite{Grad-CAM}, and attention branch network (ABN)\cite{ABN}. 
CAM first applies global average pooling (GAP) \cite{GAP} to the feature maps acquired by convolution. 
It then uses the average value of the feature maps on each channel output by GAP as weights. 
It finally generates an attention map from the weighted sum of each feature map. 
As shown in Figure \ref{fig:abn}, the ABN applies a $1\times1$ convolution to the feature map immediately before applying GAP to generate an attention map. 
Moreover, an attention mechanism that weights the feature maps on the grounds of the acquired attention maps is introduced in a manner that simultaneously improves recognition performance. 

Visual explanation can visually interpret the grounds of a decision of deep learning models by using an attention map. However, if training is insufficient owing to insufficient training data, bias, or label noise caused by annotation errors, the appropriate gaze regions are not generated. To address this problem, a method for manually correcting the attention maps of samples misrecognized by trained ABN and relearning them as teaching data was proposed by Mitsuhara et al \cite{re_teach}. 
As for this method, an expert in image-classification tasks modifies the attention map by applying their own knowledge and retrains the network using the modified attention map. 
In this way, the network can learn the appropriate gaze regions and import expert knowledge. In the case of a detailed-image identification task, the expert's knowledge was used to modify the attention map so that it focuses on more-detailed regions of the object. 
As a result, recognition performance and visual explanatory power were both improved.

Visual explanation is the presentation of the grounds for the AI's judgment from the AI to the human, and the incorporation of expert knowledge in the ABN used by the AI is an effort by the human to intervene in the AI's learning. 
On the contrary, this study aims to have AI intervene in a human's learning. As for the proposed method, the attention map is used as a cue to connect the human and AI, and the learner edits the attention map while checking changes in the recognition results of the ABN (which incorporates the expert's knowledge). 
Repeating this interactive operation makes it possible to learn the grounds of the judgments of the expert embedded in the AI in a manner that is more effective than normal learning.

\begin{figure}[tb]
\centering
\includegraphics[width=12.2cm]{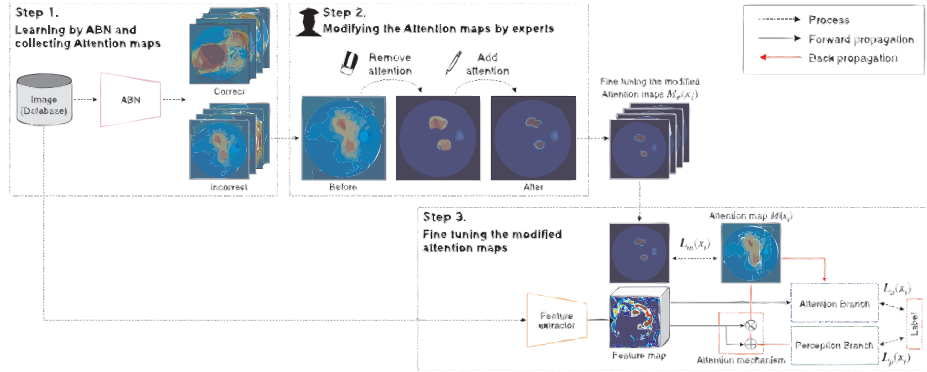}
\caption{Flow of incorporation of expert knowledge}
\label{fig:emb_expert_knowledge}
\end{figure}

\section{Proposed method}
As for the proposed learning method, the learner interactively learns the grounds of the decision by using an ABN, which incorporates the expert's knowledge, as a teacher. 
First, expert knowledge is imported into the ABN to create a teaching AI model. 
The learner then edits the attention map and applies the edited map to the attention mechanism of the ABN to output classification results, which the learner then learns from the AI. For this learning, we created an educational application for editing attention maps and making inferences. 
Using the created application, the learner can learn the exact gaze region by editing the attention map interactively in a manner that allows novice learners with no expertise to learn the expert's knowledge on their own. 
In this study, diabetic retinopathy, an ocular fundus disease, was the subject of judging the presence of disease in fundus images.

\subsection{Creating an AI model that incorporates knowledge of experts}
First, expert knowledge is incorporated into the ABN to create the teaching AI. 
The flow of incorporating expert knowledge into ABN is shown in Figure \ref{fig:emb_expert_knowledge}. Incorporating the expert knowledge consists of the following three steps: 

\noindent
\textbf{Step 1. }
Use the learned ABN to collect misidentified training samples and corresponding attention maps.

\noindent
\textbf{Step 2. }
The attention map of a misidentified sample is modified on the grounds of the expert's knowledge so that the gaze is on the region of the sample that is the grounds for the judgment of disease or no disease. 
$\mathcal{D}$ is taken as the set of training samples, and $x_i \in \mathcal{D}$. 
Also $M_e(x_i)$ is taken as the attention map of $x_i$ as modified by the expert.

\noindent
\textbf{Step 3. }
The modified attention map $M_e(x_i)$ is used for fine tuning the ABN, where learning error $L(x_i)$ is defined as
\begin{equation}
    L(x_i) = L_a(x_i) + L_p(x_i) + \lambda L_m(x_i)
\end{equation}
where $\lambda$ is a scale parameter for $L_m(x_i)$, and $L_a(x_i)$ and $L_p(x_i)$ are the training errors (obtained from cross-entropy errors) for the classification results from the attention branch and perception branch, respectively, which are used in the usual training of the ABN. 
In contrast, $L_m$ is the error when outputting the same attention map $M_e(x_i)$ as the corrected attention map $M(x_i)$, and it is defined as
\begin{equation}
    L_m(x_i) = \left\| M_e(x_i) - M(x_i) \right\|_2
\end{equation}
Note that only the parameters of the attention and perception branches are updated when fine tuning the modified attention maps, not the entire ABN network.

As described above, incorporating expert knowledge into the ABN makes it possible to focus on the same regions as experts and output correct classification results.

\subsection{Inference processing using edited attention maps}
As for the proposed learning method, the learner learns by editing the attention maps and checking the classification results of the ABN by referring to the edited attention maps. 
Hereafter, the inference process of the ABN using the attention maps edited by the learner is described. 

 The attention map edited by the learner for training sample $x_i$ is taken as $M'(x_i)$. 
 Note that the elements of a normal attention map output by an ABN are continuous values $[0, 1]$, whereas each element of the edited attention map is a binary value $\{0, 1\}$. 
 The edited attention map $M'(x_i)$ is utilized in the attention mechanism of the ABN in a manner that outputs classification results that take into account the regions focused on by the learner. 
 If $g(x_i)$ is taken as the feature map output from the feature extractor of the ABN, feature map $g'(x_i)$ weighted by using $M'(x_i)$ is defined as
\begin{equation}
    g' (x_i) = g (x_i) \cdot \left(1 + M'(x_i) \right)
\end{equation}
The weighted feature map $g'(x_i)$ is input to the perception branch, which outputs the classification result when the regions of the edited attention map are focused on.

In this way, the learner can edit the attention map and see the corresponding classification results. 
By repeatedly editing the attention maps and checking the scores of the classification results, the learner learns the relationship between the fluctuation of the classification results and the corresponding gaze regions, and they can learn the appropriate gaze regions (namely, the knowledge of the experts) by themselves.

\begin{figure}[tb]
\centering
\includegraphics[width=12.2cm]{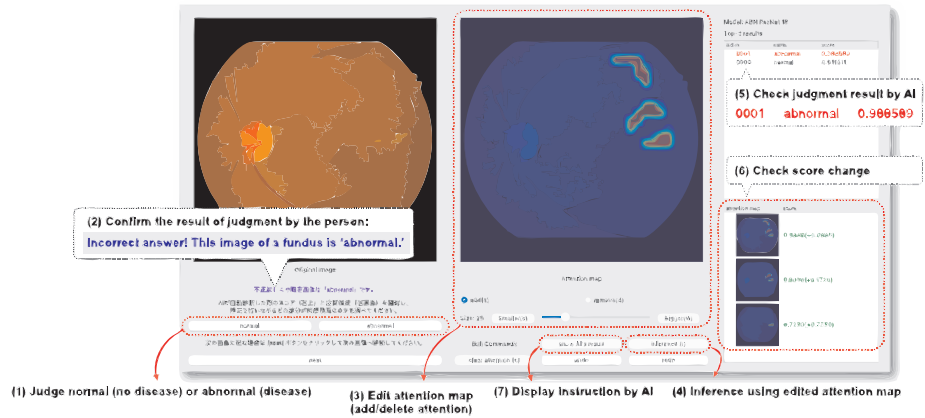}
\caption{Overview of educational application}
\label{fig:app_image}
\end{figure}

\begin{figure}[tb]
\centering
\includegraphics[width=12.2cm]{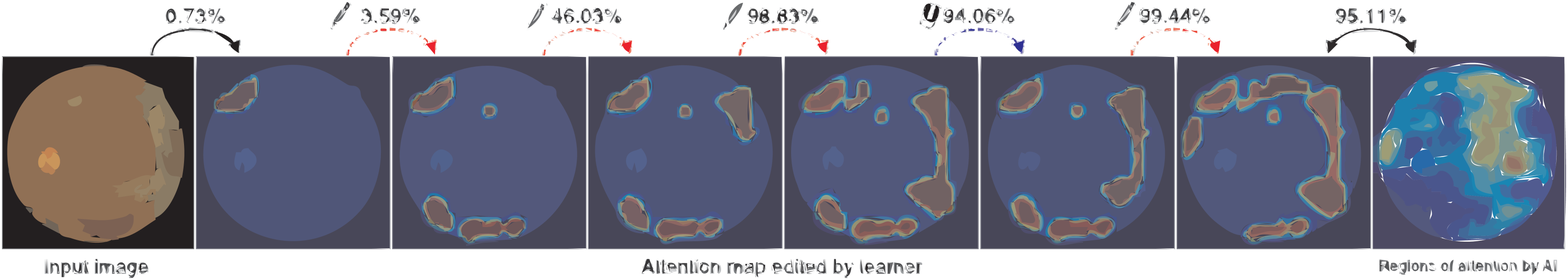}
\includegraphics[width=12.2cm]{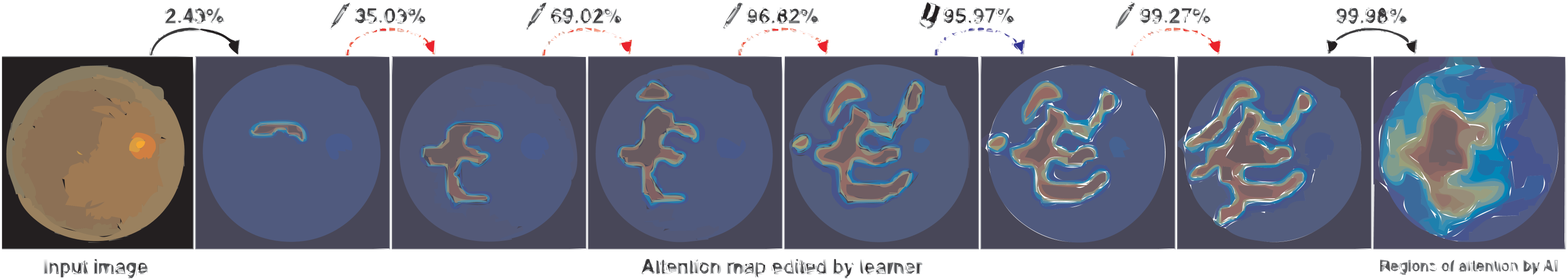}
\caption{Editing of attention maps by using educational app and changes in classification results}
\label{fig:score_trans}
\end{figure}

\subsection{“Learning from AI” educational application}
To effectively implement the proposed learning method, we created an educational application (“app” hereafter). 
The app aims to enable interactive learning between the learner and the AI by allowing the learner to easily operate the aforementioned functions. 
The educational app based on the proposed learning method is overviewed in Figure \ref{fig:app_image}, and the operation procedure of the app is explained below.

\noindent
\textbf{Step 1. }
A randomly selected sample from the training dataset is displayed, and the learner judges the classification (“normal” or “diseased”).

\noindent
\textbf{Step 2. }
Check whether the selected judgment result is correct. 

\noindent
\textbf{Step 3. }
(If the judgment in Step 2 is incorrect.) The learner edits the attention map so that it matches the correct-judgment label and emphasizes the regions that could be the grounds for the judgment.

\noindent
\textbf{Step 4. }
Using the edited attention map, the learner makes inferences with ABN.

\noindent
\textbf{Step 5. }
Confirm the classification result by the ABN.

\noindent
\textbf{Step 6. }
Repeat Steps 3 to 5. During this repetition, considering the changes in the scores of the classification results, the learner learns the gaze regions in which the scores increase by comparing their own edited attention map with that of the ABN over time.

\noindent
\textbf{Step 7. }
After Steps 3 to 6 are completed, the attention maps of the ABN are displayed as study samples. 
The learner then compares their edited attention map with the one shown by the ABN; accordingly, they learn which regions to pay attention to. 

\noindent
\textbf{Step 8. }
Return to Step 1 and repeat the training with different samples.

By repeating the process from Step 3 to Step 6, the learner learns interactively with the AI. 
If the learner responds correctly to the label, Steps 3 to 6 are unnecessary, and training continues with a different sample. Moreover, by following Step 7, the learner is expected to learn the grounds of the judgment based on the expert's knowledge by checking the attention map output by the ABN. 
Two examples of editing an attention map by using the educational app and the corresponding changes in classification results are shown in Figure \ref{fig:score_trans}. 
It can be seen that adding appropriate regions increases the score, and adding inappropriate regions decreases the score; thus, active learning is expected to be efficient and highly effective.

\section{Experimental results}
The effectiveness of the proposed teaching method was verified by an evaluation experiment using human subjects. 
In the experiment, to mitigate differences in judgment ability due to age, 36 subjects in their 20s or so were used. 
Moreover, people with neither health problems nor prior knowledge were recruited as subjects by means of a questionnaire on their health status and prior knowledge at the time of participation in the experiment. Before the experiment, the 36 subjects were divided into the following three groups. 
The first group was trained without using the learning app, the second group was trained with an AI model without expert knowledge, and the third group was trained with an AI model incorporating expert knowledge. 
The results obtained by these three different learning methods used by the three groups confirmed the effectiveness of the proposed learning method.

\subsection{Experimental procedure}
First, as a preliminary test before the start of the learning process, 60 images were presented to the subjects, who judged the presence or absence of disease. 
A time limit was not imposed on the subjects when responding to the images. 
According to the percentage of correct answers in this preliminary test, the subjects were divided into three groups with similar averages. The subjects then learn by using the learning method assigned to their group. 
The learning time was set to 30 minutes for all groups. 
After the learning was completed, a test was conducted in the same setting as the preliminary test. Using the same setting makes it is possible to determine if samples that were misclassified before learning were correctly classified by using each learning method. 
Each learning method is described as follows.



\begin{table*}[t] 
\centering
\caption{Correct answer rate achieved by each subject (\%)} 
\label{table:teach_table}
\tabcolsep = 0.8mm
\begin{tabular}{c|cc|cc|cc}
\hline
& & & \multicolumn{4}{|c}{Proposed method} \\\cline{4-7}
& \multicolumn{2}{c}{No education app} & \multicolumn{2}{|c}{AI model without} & \multicolumn{2}{|c}{AI model with}\\
& \multicolumn{2}{c}{ } & \multicolumn{2}{|c}{knowledge} & \multicolumn{2}{|c}{knowledge}\\\cline{2-7}
& Before & After   & Before  & After   & Before  &After\\
& learning & learning & learning & learning & learning & learning\\
\hline
\multirow{12}{*}{Subject} & 56.67 & 70.00 & 55.00 & 51.67 & 50.00 & 75.00\\
 & 48.33 & 76.67 & 63.33 & 56.67 & 50.00 & 75.00\\
 & 46.67 & 78.33 & 43.33 & 60.00 & 45.00 & 78.33\\
 & 61.67 & 76.67 & 41.67 & 66.67 & 63.33 & 78.33\\
 & 43.33 & 75.00 & 60.00 & 70.00 & 38.33 & 78.33\\
 & 43.33 & 76.67 & 71.67 & 73.33 & 55.00 & 80.00\\
 & 38.33 & 78.33 & 33.33 & 73.33 & 56.67 & 80.00\\
 & 68.33 & 78.33 & 53.33 & 75.00 & 51.67 & 81.67\\
 & 73.33 & 76.67 & 66.67 & 76.67 & 56.67 & 81.67\\
 & 70.00 & 80.00 & 68.33 & 76.67 & 58.33 & 81.67\\
 & 53.33 & 81.67 & 58.33 & 78.33 & 41.67 & 81.67\\
 & 65.00 & 85.00 & 75.00 & 80.00 & 75.00 & 81.67\\\hline
Average & 55.69$\pm$11.86 & 77.78$\pm$3.65 &  57.50$\pm$12.82  & 69.86$\pm$9.20 & 53.47$\pm$9.88 & \textbf{79.44}$\pm$2.50 \\
\hline
\end{tabular}
\end{table*}

\textbf{Learning without the app} 
The educational app does not provide interactive learning; instead, it presents information about the fundus images and their correct labels, and the subject learns from that information alone.

\textbf{Learning using the app with AI model incorporating expert knowledge} 
The app is used for learning, and an AI model with no expert knowledge is used as the teacher. 
Here, an AI model with no expert knowledge means an ABN obtained only by ordinary supervised learning using training data and correct labels.

\textbf{Learning using the app with AI model incorporating expert knowledge} 
The AI model and educational app described in Section 3 are used for learning.

\begin{figure}[tb]
\centering
\includegraphics[width=12.2cm]{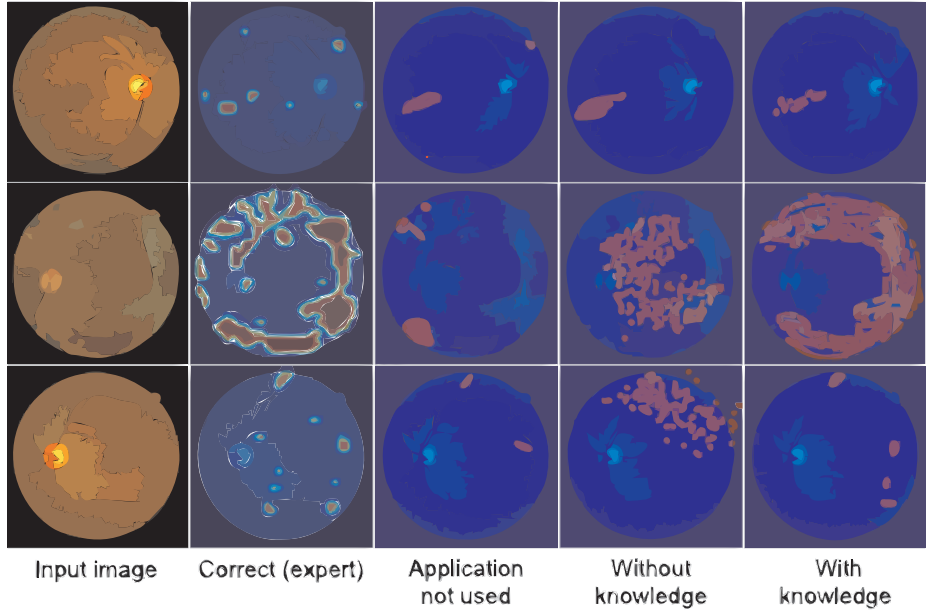}
\caption{Comparison of attention maps}
\label{fig:ans_img}
\end{figure}

\subsection{Datasets}
The effectiveness of learning by using the proposed method was experimentally evaluated on the basis of the percentage of correct responses in binary classification for judgment of disease in fundus images. 
The effectiveness of the proposed method could be appropriately evaluated because the images are difficult to classify accurately without expert knowledge and cannot be observed on a daily basis.

As datasets of fundus-disease images, the Messidor Dataset \cite{Messidor} for diabetic retinopathy grading and the Indian Diabetic Retinopathy Image Dataset (IDRiD) \cite{IDRiD} were used. 
The Messidor Dataset has four levels of grading, from “0” to “3,” with the larger values indicating greater severity of disease. 
For the pre-test and post-test conducted by the subjects, 30 grade-0 images and 10 images each of grades 1 to 3 were taken from the Messidor Dataset. 
The IDRiD dataset was used for learning; in particular, 124 normal-fundus images and 81 diseased-fundus images were used as the learning materials. 
Classification accuracy of the model for the test with the IDRiD dataset was 94.44\% for the model without expert knowledge and 97.22\% for the model incorporating expert knowledge.

\subsection{Evaluation by classification accuracy}
Classification accuracies of each group are listed in Table \ref{table:teach_table}. 
According to the results in the table, compared with the other learning methods, the proposed method using the AI model with expert knowledge results in higher post-learning scores. 
Moreover, there is large variation in scores of the group that did not use the app and the group that used the app with the AI model without expert knowledge. 
This variation might be due to the fact that learning and subsequent solving of the problem (judgment of disease or no disease) is dependent on individual ability. 
On the contrary, the groups that learned with the proposed method using the AI model with expert knowledge showed less variation, and that trend indicates that the proposed method is effective for teaching multiple learners.

\begin{table*}[t] 
\centering
\caption{Class IoU scores for attention maps of subjects} 
\label{table:ans_table}
\tabcolsep = 3.0mm
\begin{tabular}{c|cc|cc|cc}
\hline
& & & \multicolumn{4}{|c}{Proposed method} \\\cline{4 - 7}
& \multicolumn{2}{c}{No education app} & \multicolumn{2}{|c}{AI model without} & \multicolumn{2}{|c}{AI model with}\\
& \multicolumn{2}{c}{ } & \multicolumn{2}{|c}{knowledge} & \multicolumn{2}{|c}{knowledge}\\\cline{2-7}
& Before & After   & Before  & After   & Before  &After\\
& learning & learning & learning & learning & learning & learning\\ 
\hline
\multirow{4}{*}{Subject} 
     & 0.0974 & 0.1194 & 0.0728 & 0.1346 & 0.1083 & 0.1931 \\
     & 0.1479 & 0.1378 & 0.0740 & 0.0670 & 0.0983 & 0.0949 \\
     & 0.0314 & 0.0983 & 0.1090 & 0.0659 & 0.1782 & 0.1262 \\
     & 0.0983 & 0.0762 & 0.0938 & 0.0811 & 0.0555 & 0.0761 \\\hline
Average & 0.0958 & 0.1079 & 0.0895 & 0.0835 & 0.1107 & \textbf{0.1314} \\
\hline
\end{tabular}
\end{table*}

\subsection{Evaluation by attention maps}
Next, the attention maps edited by the subjects were compared so as to verify the learning effect of the developed app based on the proposed method, namely, whether the subjects are paying attention to the appropriate regions. 
Examples of attention maps compiled by a subject in each group are shown in Figure \ref{fig:ans_img}. It is clear from the figure that the subject who learnt with the proposed method compiled attention maps similar to those of the expert. 
Although the subject using the proposed method using expert knowledge added some parts to match the expert's attention, the subject did not add other regions of attention.

As for a quantitative evaluation using the attention maps, the class intersection-over-union (IoU) scores for the disease regions of the attention maps for the 30 disease images of the subjects are listed in Table \ref{table:ans_table}. 
The proposed method with the expert-knowledge AI model achieves the highest class IoU after learning, and that result indicates that it enables the subject to compile an attention map that is closest to that of the expert. 
The above results demonstrate that the subject interactively learned which regions to focus on by learning with the AI model incorporating the knowledge of an expert. 
They also demonstrate that a highly effective learning can be achieved by showing the grounds of judgment of disease to the learner by means of the attention map shown by the AI.

\subsection{Feedback from subjects}
As a qualitative evaluation of the developed educational application, feedback from the subjects after they used the application to learn using the AI model incorporating expert knowledge is listed below:

\begin{itemize}
    \item “I think I could learn actively.”
    \item “I think it is an appropriate educational application.”
    \item “Just learning how to use the educational application helped me understand the essence of the disease.”
\end{itemize}
In contrast, feedback from the subjects trained using the AI model incorporating no expert knowledge is listed below:

\begin{itemize}
    \item “The results of editing suspected disease regions were not linked to evaluation by AI.”
    \item“I failed to focus on detailed regions.”
    \item “It was difficult to correct my assumptions about the characteristics of the disease because the focus was on regions differing from the ones that I had judged as disease.”
\end{itemize}
These results demonstrate that the educational application with the AI model that incorporates expert knowledge clarifies the regions of disease as a grounds for judgment and facilitates the learning effect of the proposed method.

\section{Concluding remarks}
A method that enables a learner to “learns from AI” by means of a deep-learning model incorporating expert knowledge as a teacher was proposed. In detail, the attention maps and attention mechanism of an attention branch network (ABN) are applied in a manner that imports expert knowledge into the network. 
Using the ABN incorporating expert knowledge, the learner can edit the attention maps and check the classification results by applying the map to the attention mechanism; as a result, the leaner can learn on the basis of the judgment grounds of the AI model. 
To implement the proposed learning method, an educational application was created and used to evaluate the method in experiments with subjects. 
The results of the experiment using images of ocular fundus disease showed that the proposed learning method was more effective than the standard method and enabled the subjects (learners) to focus on the same disease regions as the expert when judging the images.

As for future work, we hope to perform larger-scale subject experiments, improve the application, and apply the proposed “learn from AI” method to tasks that require a higher level of expertise (such as visual inspection) in addition to judgment of disease in medical images as described in this report.

\section{ACKNOWLEDGMENTS}
This paper is based on results obtained from a project, JPNP18002, commissioned by the New Energy and Industrial Technology Development Organization (NEDO).

\bibliographystyle{unsrt}
\bibliography{ref}

\end{document}